\documentclass[11pt,a4paper]{article}

\usepackage{times,latexsym}
\usepackage{url}
\usepackage[T1]{fontenc}

\usepackage{textcomp}
\usepackage[utf8]{inputenc}
\usepackage{float}
\usepackage{microtype}

\usepackage{inconsolata}
\usepackage[fleqn]{amsmath}
\usepackage{graphicx}
\usepackage{pgfplots, pgfplotstable}
\usepackage{amssymb}
\usepackage{algpseudocode}
\usepackage{algorithm}
\usepackage{mathalpha}
\usepackage{mathtools}
\usepackage{relsize}
\usepackage{tabularx}
\usepackage{siunitx}

\usepackage{booktabs}

\usepackage[russian,USenglish]{babel}
\MakeRobust{\Call}

\DeclareMathOperator*{\argmax}{argmax} 
\usepackage{multirow}
\usepackage[flushleft]{threeparttable}
\newcommand{\RNum}[1]{\lowercase\expandafter{\romannumeral #1\relax}}

\usepackage{substitutefont}
\substitutefont{T2A}{\rmdefault}{Tempora-TLF}

\pgfplotsset{compat=1.17} 
\pgfplotstableread[col sep=&, header=true]{
	firm    &value1   &value2  &value3
	4	&31.10	&30.46	&31.05
	5	&31.09	&30.64	&31.03
	6	&31.11	&30.57	&31.08
	7	&31.10	&30.68	&31.07
	8	&31.12	&30.79	&31.04
	9	&31.15	&30.76	&31.10
	10	&31.16	&30.69	&31.14
	20	&31.20	&30.73	&31.18
	30	&31.19	&30.79	&31.16
	40	&31.18	&30.76	&31.15
	50	&31.20	&30.83	&31.17
	60	&31.00	&30.69	&30.98
	70	&31.20	&30.70	&31.16
	80	&31.21	&30.72	&31.18
	90	&31.23	&30.72	&31.19
}\RerankingResults

\pgfplotsset{compat=1.17} 
\pgfplotstableread[col sep=&, header=true]{
    nbest	  &BaseDoc &TransCoref  
    4	      &34.03 	&34.26 
    5	      &34.89 	&35.00 
    6	      &35.36 	&35.50 
    7	      &35.78 	&36.03 
    8	      &36.18 	&36.34 
    9	      &36.49 	&36.67 
    10	      &36.85 	&36.87 
    20	      &38.62 	&38.93 
    30	      &39.48 	&39.69 
    40	      &40.16 	&40.34 
    50	      &40.58 	&40.92 
    60	      &40.98 	&41.33 
    70	      &41.31 	&41.69 
    80	      &41.58 	&41.98 
    90	      &41.83 	&42.21 
}\OracleResults

\pgfplotsset{compat=1.17} 
\pgfplotstableread[col sep=&, header=true]{
firm    &TransCorefRR   &BaseDoc  &TransCoref &BaseDocOracle &TransCorefOracle
4	&31.10	&30.46	&31.05	  &34.03 	&34.26 
5	&31.09	&30.64	&31.03	  &34.89 	&35.00 
6	&31.11	&30.57	&31.08	  &35.36 	&35.50 
7	&31.10	&30.68	&31.07	  &35.78 	&36.03 
8	&31.12	&30.79	&31.04	  &36.18 	&36.34 
9	&31.15	&30.76	&31.10	  &36.49 	&36.67 
10	&31.16	&30.69	&31.14	  &36.85 	&36.87 
20	&31.20	&30.73	&31.18	  &38.62 	&38.93 
30	&31.19	&30.79	&31.16	  &39.48 	&39.69 
40	&31.18	&30.76	&31.15	  &40.16 	&40.34 
50	&31.20	&30.83	&31.17	  &40.58 	&40.92 
60	&31.00	&30.69	&30.98	  &40.98 	&41.33 
70	&31.20	&30.70	&31.16	  &41.31 	&41.69 
80	&31.21	&30.72	&31.18	  &41.58 	&41.98 
90	&31.23	&30.72	&31.19	  &41.83 	&42.21 
}\AllRerankingResults

\newcommand{\None}{\diamondsuit}
\newcommand{\Binary}{\heartsuit}
\newcommand{\Avoided}{\clubsuit}
\newcommand{\Multiple}{\spadesuit}

\usepackage[acceptedWithA]{tacl2021v1}

\usepackage{xspace,mfirstuc,tabulary}

\newif\iftaclinstructions
\taclinstructionsfalse 
\iftaclinstructions
\renewcommand{\confidential}{}
\renewcommand{\anonsubtext}{(No author info supplied here, for consistency with
TACL-submission anonymization requirements)}
\newcommand{\instr}
\fi

\iftaclpubformat 

\else

\fi

\title{Context-Aware Machine Translation with Source Coreference Explanation}

\author{
  Huy Hien Vu 
  \qquad
  Hidetaka Kamigaito
  \qquad
  Taro Watanabe
  \ \\
  Nara Institute of Science and Technology, Japan.
  \\
  \texttt{ \{vu.huy\_hien.va9, kamigaito.h, taro\}@is.naist.jp}
}

\date{}

\begin{document}
\maketitle
\begin{abstract}
Despite significant improvements in enhancing the quality of translation, context-aware machine translation (MT) models underperform in many cases. One of the main reasons is that they fail to utilize the correct features from context when the context is too long or their models are overly complex. This can lead to the explain-away effect, wherein the models only consider features easier to explain predictions, resulting in inaccurate translations. To address this issue, we propose a model that explains the decisions made for translation by predicting coreference features in the input. We construct a model for input coreference by exploiting contextual features from both the input and translation output representations on top of an existing MT model. 
We evaluate and analyze our method in the WMT document-level translation task of English-German dataset, the English-Russian dataset, and the multilingual TED talk dataset, demonstrating an improvement of over 1.0 BLEU score when compared with other context-aware models.
\end{abstract}

\section{Introduction}
With the rapid development of machine learning techniques, the Machine Translation (MT) field has witnessed changes from exclusively probabilistic models \citep{Brown90,Koehn2003}  to neural network based models, such as simplistic Recurrent Neural Network (RNN) based encoder-decoder models \citep{Sutskever14} or higher-level attention-based models \citep{Bahdanau14, luong2015}, and finally turn to the current state-of-the-art Transformer model \citep{Vaswani17} and its variations.

The quality of MT models, including RNN-based, attention-based, and Transformer models, has been improved by incorporating contextual information (\citealp{voita2018, wang2017}; and others.), or linguistic knowledge (\citealp{bugliarello2020, sennrich16}; and others). In the former context-aware methods, many successful approaches focus on context selection from previous sentences \cite{jean17, wang2017} using multiple steps of translation, including additional module to refine translations produced by context-agnostic MT system, to utilize contextual information \citep{voita19,xiong19}, and encoding all context information as end-to-end frameworks \citep{zhang20, bao21}. Although they have demonstrated improved performance, there are still many cases in which their models perform incorrectly for handling, i.e., the ellipsis phenomenon in a long paragraph.  
One of the reasons is that their models are still unable to select the right features from context when the context is long, or the model is overly complex. Therefore, the model will easily suffer from an explain-away effect \citep{Klein2002, Yu2017, Shah2020, Krause2023} in which a model is learned to use only features which are easily exploited for prediction by discarding most of the input features.

In order to resolve the problem of selecting the right context features in the context-aware MT, we propose a model which \textit{explains decisions of translation by predicting input features}. The input prediction model employs the representations of translation outputs as additional features to predict contextual features in the inputs. In this work, we employ coreference as the prediction task since it captures the relation of mentions that are necessary for the context-aware model. The prediction model is constructed on top of an existing MT model without modification in the same manner as done in multi-task learning, but it fuses information from representations used for the decisions of translation in the MT model.

Under the same settings of the English-Russian (En-Ru) dataset and the WMT document-level translation task of the English-German (En-De) dataset, our proposed technique outperforms the standard transformer-based neural machine translation (NMT) model in both sentence and context-aware models, as well as the state-of-the-art context-aware model measured by BLEU \citep{post2018}, BARTScore \citep{yuan21} and COMET \citep{rei2020}, and  the human-annotated test set in a paragraph \citep{voita19}. Additionally,  in the multilingual experiments, our method shows consistent results, paralleling those in the En-Ru and En-De datasets, and proving its versatility across languages.

Further analysis shows that our  coreference explanation sub-model consistently enhances the quality of translation, regardless of type of dataset size. Notably, the model demonstrates consistent improvement when additional context is incorporated, highlighting its effectiveness in handling larger context sizes. 
Additionally, the analysis highlights a strong correlation between the self-attention heat map and coreference clusters, underscoring the significance of our coreference prediction sub-model in capturing coreference information during the translation process. Moreover, our proposed training method proves to be effective in the coreference prediction task. We also provide a suggestion to finetune the contribution of the sub-model to optimize its impact within the overall MT system. We release our code and hyperparameters at \href{https://github.com/hienvuhuy/TransCOREF}{\textit{https://github.com/hienvuhuy/TransCOREF.}} 
\section{Backgrounds}
\label{sec_background}
\subsection{Transformer-based NMT}
\label{sec_background_transformernmt}
Given an input single sentence $\boldsymbol{x} = (x_1, ..., x_{|\boldsymbol{x}|})$ and its corresponding translation $\boldsymbol{y} = (y_1, ..., y_{|\boldsymbol{y}|})$, an MT system directly models the translation probability
\begin{align}
    p(\boldsymbol{y}| \boldsymbol{x}; \theta) &=  \prod_{t=1}^{|\boldsymbol{y}|}{p(y_t|\boldsymbol{y}_{<t},\boldsymbol{x}; \theta)},
\end{align}
where $t$ is the index of target tokens, $ \boldsymbol{y}_{<t}$ is the partial translation before $y_t$, and $\theta$ is the model parameter. 
At inference time, the model will find the most likely translation $\boldsymbol{\hat{y}}$ for a given source input
\begin{align}
    \boldsymbol{\hat{y}} =  \argmax_{\boldsymbol{y}} \prod_{t=1}^{|\boldsymbol{y}|}p(y_t| \boldsymbol{y}_{<t}, \boldsymbol{x}; \theta).
\end{align}
To model the translation conditional probability $p(\boldsymbol{y}|\boldsymbol{x}; \theta)$, many encoder-decoder architectures have been proposed based either on CNNs \citep{Gehring17} or self-attention \citep{Vaswani17}, and we focus on the Transformer \citep{Vaswani17} as our building block, given its superior ability to model long-term dependencies and capture context information features.
The encoder of the Transformer comprises $l_{e}$ stacked layers which transforms the input $\boldsymbol{x}$ into hidden representations $\mathbf{H}^{l_e}_{enc} \in \mathbb{R}^{|\boldsymbol{x}| \times d}$ where $d$ is a dimension for hidden vector representation. Similarly, the decoder of the Transformer comprises $l_d$ stacked layers which consumes the translation prefix $\boldsymbol{y}_{<t}$ and $\mathbf{H}^{l_e}_{enc}$ to yield the final representation $\mathbf{H}^{l_d}_{dec} \in \mathbb{R}^{|\boldsymbol{y}| \times d}$. The two processes can be formally denoted as
\begin{align}
    & \mathbf{H}_{enc}^i = \textsc{Enc}(\mathbf{H}_{enc}^{i-1})\\
    & \mathbf{H}_{dec}^i = \textsc{Dec}(\mathbf{H}_{dec}^{i - 1}, \mathbf{H}_{enc}^{l_e}).
\end{align}
Note that $\mathbf{H}^{0}_{enc}$ is the representation of $\boldsymbol{x}$ from the embedding layer, and $\mathbf{H}^{0}_{dec}$ is the representation of  $\boldsymbol{y}$ after embedding lookup with shifting by the begin-of-sentence token. $\textsc{Enc}(\cdot)$ and $\textsc{Dec}(\cdot)$ denote the function of the single Transformer encoder and decoder layer, respectively. 

The output target sequence is predicted based on the output hidden state $\mathbf{H}_{dec}^{l_d}$ from the top layer of the decoder
\begin{align}
    p(y_t|&\boldsymbol{y}_{<t},\boldsymbol{x}; \theta)\nonumber\\
    &=         \textsc{Softmax}(\mathbf{W}_{dec}\mathbf{H}_{dec}^{l_d}[t])[y_t]         
\end{align}
where $\mathbf{W}_{dec} \in \mathbb{R}^{|\mathcal{V}| \times d}$ is the projection weight matrix which maps the hidden state to the probability in the output vocabulary space $\mathcal{V}$, and $[\cdot]$ denotes an index/slice to a vector/matrix.

The standard training objective is to minimize the cross-entropy loss function
\begin{align}
    \label{loss_cross_entropy}	& \mathcal{L}_{\textsc{MT}} = -\sum_{ (\boldsymbol{x}, \boldsymbol{y}) \in \mathcal{D}} \sum_{t=1}^{|\boldsymbol{y}|} \log p(y_t| \boldsymbol{y}_{<t}, \boldsymbol{x};\theta)
\end{align}
given a parallel corpus $\mathcal{D}=\{(\boldsymbol{x}^w, \boldsymbol{y}^w) \}_{w=1}^{|\mathcal{D}|}$ which contains $|\mathcal{D}|$ pairs of single sentence and its corresponding translation.

\subsection{Context-Aware Transformer-base NMT}
\label{sec_context_aware_tranformer_nmt}
A context-aware MT model can be regarded as a model which takes a document, i.e., multiple sentences, as an input and generates multiple sentences as its corresponding translation.
We assume that each sentence is translated into a single sentence, and define the source document $\boldsymbol{\underline{x}} = (\boldsymbol{x}^1,...,\boldsymbol{x}^{n})$ with $n$ sentences and its corresponding target language document $\boldsymbol{\underline{y}} = (\boldsymbol{y}^1,...,\boldsymbol{y}^{n})$. A context-aware MT system directly models the translation probability
\begin{align}
    p(\boldsymbol{\underline{y}}|\boldsymbol{\underline{x}}; \theta)  &=		 \sum_{k=1}^{n}{p(\boldsymbol{y}^k|\boldsymbol{y}^{<k},\boldsymbol{\underline{x}} ; \theta)},
\end{align}
where $k$ is an index to a sentence in $\boldsymbol{\underline{y}}$, $ \boldsymbol{y}^{<k}$ is the partial translation before the sentence $\boldsymbol{y}^k$. In this model, we assume that $ \left\langle \boldsymbol{\underline{x}},\boldsymbol{\underline{y}} \right\rangle $ constitute a parallel document and each $\left\langle \boldsymbol{x}^k,\boldsymbol{y}^k \right\rangle$ forms a parallel sentence.

Several approaches can be used to produce a translated document, i.e., keeping a sliding window of size $m$ \citep{tiedemann17}, joining these $m$ sentences as a single input, translating these $m$ sentences and selecting the last sentence as an output ($m$-to-$m$) \citep{zhang20}, or joining whole sentences in a document as a very long sequence and translating this sequence \citep{bao21}, amongst other methods. To simplify the definition of the context-aware NMT model, we opt for the $m$-to-$m$ method and use a special character (\textit{\_eos}) between sentences when feeding these $m$ sentences to the model. In this way, the context-aware translation model can still be defined as a standard sentence-wise translation in \S \ref{sec_background_transformernmt}.

\subsection{Coreference Resolution task}
\label{sec_coref-task}
Coreference Resolution is the task of identifying and grouping all the mentions or references of a particular entity within a given text into a cluster, i.e., a set of spans. This task has progressed significantly from its earlier approaches, which were based on hand-crafted feature systems \cite{mccarthy95, aone1995}, to more advanced and effective deep learning approaches based on spans-ranking \cite{lee2017, lee2018, kirstain21} and for multilingual languages \cite{zheng2023multilingual}.

It is typically formulated as explicitly identifying an antecedent span to the left of a mention span in the same cluster. More formally, a set of clusters $\boldsymbol{\mathcal{C}} = \{..., \mathcal{C}_k, ...\}$ is predicted for an input sequence $\boldsymbol{x}$, either a document or a single sentence, with each cluster comprising a set of non-overlapping spans $\mathcal{C}_k = \{(i, j) : 1 \leq i \leq j \leq |\boldsymbol{x}|\}$ ($1 \leq k \leq |\boldsymbol{\mathcal{C}}|$). We introduce an alternative view using a variable $\mathcal{A}$ which represents mapping for all possible mention spans $\mathcal{S} = \{(i, j): \forall 1 \leq i \leq j \leq |\boldsymbol{x}|\}$ of $\boldsymbol{x}$ to its antecedent span within the sample cluster $\mathcal{C}_k$, i.e., $\mathcal{A} = \{..., s \rightarrow c, ...\}$, where $c \in \mathcal{C}_k$ is an antecedent to the left of $s \in \mathcal{C}_k$ and $c = \epsilon$, i.e., an empty span, when $s$ is not a member of any clusters $\mathcal{C}_k$. Note that we can derive a unique $\boldsymbol{\mathcal{C}}$ given a single derivation of $\mathcal{A}$ by forming a cluster of spans connected by antecedent links, but there are multiple derivations of $\mathcal{A}$ for $\boldsymbol{\mathcal{C}}$ when there exists a cluster $|\mathcal{C}_k| > 2$.
The task is modeled by the conditional probability distribution of independently predicting any possible antecedents of a mention span in the same cluster
\begin{align}
&p(\boldsymbol{\mathcal{C}} | \boldsymbol{x}) = \sum_{\mathcal{A} \in a(\boldsymbol{\mathcal{C}})} p(\mathcal{A}| \boldsymbol{x})\nonumber\\
&= \prod_{s \in \mathcal{S}} \sum_{\mathcal{A} \in a(\boldsymbol{\mathcal{C}})} p(\mathcal{A}_s | s, \boldsymbol{x})\nonumber\\
& = \prod_{s \in \mathcal{S}} \sum_{\mathcal{A} \in a(\boldsymbol{\mathcal{C}})} \frac{\exp\left(f(\mathcal{A}_s, s; \mathbf{H}_{coref})\right)}{\sum_{c \in \mathcal{M}_s} \exp\left(f(c, s; \mathbf{H}_{coref})\right)}\nonumber\\
& \triangleq \prod_{s \in \mathcal{S}} \sum_{\mathcal{A} \in a(\boldsymbol{\mathcal{C}})} \textsc{Coref}(\mathcal{A}_s, s; \mathbf{H}_{coref}),
\end{align}
where $a(\cdot)$ is a function that returns all possible derivations for clusters and $\mathcal{M}_s$ is a set of all possible spans to the left of $s$ including $\epsilon$.
 $f(\cdot , \cdot)$ is a score function \citep{kirstain21} to compute both mention and antecedent scores and $\mathbf{H}_{coref} \in \mathbb{R}^{|\boldsymbol{x}| \times d}$ is contextualized representation of the input sequence  $\boldsymbol{x}$, i.e., BERT \citep{devlin19}. We denote the final function as $\textsc{Coref}(\cdot, \cdot)$ for brevity.

We adopt the training scheme proposed by \citet{kirstain21}, which filters spans to avoid the explicit enumeration of all possible mention spans, and represents antecedent relations using only the endpoints of the retained spans with a biaffine transformation.
At the training stage, we minimize the negative log-likelihood of predicting clusters
\begin{align}
\label{loss_marginal_log_likelihood}
&\mathcal{L}_{\textsc{Coref}} = - \!\!\!\!\!\! \sum_{\boldsymbol{(\mathcal{C}}, \boldsymbol{x}) \in \mathcal{D}_{\textsc{Coref}}} \!\!\!\!\!\!\log \prod_{s \in \mathcal{S}} \sum_{\mathcal{A} \in a(\boldsymbol{\mathcal{C}})} p(\mathcal{A}_s | s, \boldsymbol{x}),
\end{align}
where $\mathcal{D}_{\textsc{Coref}}$ is a training data for coreference resolution.
\section{Context-Aware MT with Coreference Information}
\label{sec_proposed_method}
Our motivation stems from the observation that when translating a paragraph, translators are able to pick up precise words and explain why a particular choice of word is better given the context especially by relying on linguistic cues such as discourse structure, verb equivalence, etc. Thus, instead of modeling a translation by directly relying on an additional conditional variable of coreference clusters $\boldsymbol{\mathcal{C}}$ for $\boldsymbol{x}$, i.e., $p(\boldsymbol{y} | \boldsymbol{\mathcal{C}}, \boldsymbol{x})$, we propose a model that is akin to the noisy channel framework \cite{yee2019}, to explain the decision $\boldsymbol{y}$ made by the translation model :
\begin{flalign}
    \label{joint_models}p(\boldsymbol{y} | \boldsymbol{\mathcal{C}}, \boldsymbol{x}) &= \frac{p(\boldsymbol{y}, \boldsymbol{\mathcal{C}}| \boldsymbol{x}) }{p(\boldsymbol{\mathcal{C}}| \boldsymbol{x})}
    = \frac{p(\boldsymbol{y} | \boldsymbol{x})(\boldsymbol{\mathcal{C}} | \boldsymbol{y}, \boldsymbol{x})}{p(\boldsymbol{\mathcal{C}} | \boldsymbol{x})}\nonumber \\
    &\propto p(\boldsymbol{y} | \boldsymbol{x})p(\boldsymbol{\mathcal{C}} | \boldsymbol{y}, \boldsymbol{x}),
\end{flalign}
where $p(\boldsymbol{\mathcal{C}} | \boldsymbol{y}, \boldsymbol{x})$ is a model to predict coreference clusters given both an input sentence and its translation. Note that we can omit the denominator $p(\boldsymbol{\mathcal{C}} |  \boldsymbol{x})$  given that it is a constant when predicting $\boldsymbol{y}$, similar to the noisy channel modeling, since both $\boldsymbol{x}$ and $\boldsymbol{\mathcal{C}}$ are input to our model.  The direct model $p(\boldsymbol{y} | \boldsymbol{\mathcal{C}}, \boldsymbol{x})$ is prone to ignore features in $\boldsymbol{\mathcal{C}}$, especially when the context is long, since the information in $\boldsymbol{x}$ has direct correspondence with $\boldsymbol{y}$. In contrast, the model for coreference resolution task, $p(\boldsymbol{\mathcal{C}} | \boldsymbol{y}, \boldsymbol{x})$, will explain the coreference cluster information in  $\boldsymbol{x}$ not only by the features from  $\boldsymbol{x}$ but additional features from  $\boldsymbol{y}$ and, thus, the higher $p(\boldsymbol{\mathcal{C}} | \boldsymbol{y}, \boldsymbol{x})$, the more likely $\boldsymbol{y}$ will be a translation for  $\boldsymbol{x}$. When coupled with the translation model $p( \boldsymbol{y} | \boldsymbol{x})$ especially when jointly trained together, our formulation will be able to capture long-distance relations in coreference clusters since the coreference resolution task needs to predict it given $\boldsymbol{x}$ and $\boldsymbol{y}$.

\subparagraph{Architecture}

The two sub-models, i.e., $p(\boldsymbol{y} | \boldsymbol{x})$ and $p(\boldsymbol{\mathcal{C}} | \boldsymbol{y}, \boldsymbol{x})$, could be trained separately as done in a noisy channel modeling approach of MT \citep{yee2019}.
This work formulates it as a multi-task setting by predicting two tasks jointly, i.e., translation task and coreference resolution task, by using the representations of the encoder and decoder of \citet{Vaswani17}.
More specifically, we do not alter translation task $p(\boldsymbol{y} | \boldsymbol{x})$, but obtain the representation for the coreference task by fusing the representations of the encoder and decoder as follows
\begin{flalign}\label{fuse_representation}
\hspace{-0.5cm}p(\boldsymbol{\mathcal{C}} | \boldsymbol{y}, \boldsymbol{x}) & = \prod_{s \in \mathcal{S}} \sum_{\mathcal{A} \in a(\boldsymbol{\mathcal{C}})} \textsc{Coref}(\mathcal{A}_s, s; \mathbf{H'}_{coref}) \nonumber \\
\mathbf{H'}_{coref} & = \Call{Dec}{\boldsymbol{H}_{enc}^{l_e},\boldsymbol{H}_{dec}^{l_d}}.
\end{flalign}
Note that we obtain $\mathbf{H'}_{coref}$ from an additional decoder layer for the encoder representation $\boldsymbol{H}_{enc}^{l_e}$ with cross attention for $\boldsymbol{H}_{dec}^{l_d}$.

\subparagraph{Training}
We jointly train our two sub-models using the label-smoothing variant of the cross-entropy loss function in Equation \ref{loss_cross_entropy} and the marginal log-likelihood loss function in Equation \ref{loss_marginal_log_likelihood}, but using $\mathbf{H'}_{coref}$ in Equation \ref{fuse_representation} as follows
	\begin{flalign}
		\label{loss_fuse}	\mathcal{L} = \mathcal{L}_{\textsc{MT}} + \alpha  \mathcal{L'}_{\textsc{Coref}}  , 
	\end{flalign}
where $\alpha$ is a hyperparameter that controls a contribution of the coreference resolution task. During the training step, we feed pairs of sentences together with coreference cluster information generated by an external coreference resolution framework since human annotation is not available in MT tasks.

\subparagraph{Inference}
Inference is complex in that the model for the coreference resolution task has to be evaluated every time a target token is generated by the translation model as done in the noisy channel approach of MT \citep{yee2019}.
We resort to a simpler approach of ignoring the term for coreference clusters, i.e., $p(\boldsymbol{\mathcal{C}}| \boldsymbol{y},\boldsymbol{x})$, and using only the token prediction, i.e., $p(\boldsymbol{y}|\boldsymbol{x})$; alternatively, 
we generate a large set of $N$-best translations from $p(\boldsymbol{y}|\boldsymbol{x})$ and rerank them using the joint probabilities
\begin{flalign}\label{translation_score} \log{p(\boldsymbol{y}|\boldsymbol{x})} + \beta  \log {p(\boldsymbol{\mathcal{C}}|\boldsymbol{y}, \boldsymbol{x}}),
\end{flalign}
 where $\beta$ is a hyperparameter to control the strength of the coreference resolution task.

\section{Experiments}

\label{sec_experiments}
\subsection{Dataset}
\label{section_dataset}

We utilized the En-Ru dataset \citep{voita19} and the widely adopted En-De benchmark dataset IWSLT 2017, as used in \citet{maruf2019}, with details provided in Table \ref{dataset_overview}.
We also used the multilingual TED talk dataset \citep{qi2018} to assess the efficacy of our proposed method across a variety of language types, including different characteristics in pronouns, word order and gender assignment with specifics delineated in Table \ref{multi_lang_types}.
\begin{table}[t]\centering
\scalebox{0.9}{
	\begin{tabular}{lccc}
		\toprule
		& \begin{tabular}[c]{@{}c@{}}Avg. \#Coref. Clusters\\ train/valid/test\end{tabular} & \begin{tabular}[c]{@{}c@{}}\#Samples\\  train/valid/test\end{tabular}  \\ 
		\midrule
        En-Ru & 3.1/3.0/2.9 & 1.5M/10k/10k\\
        En-De & 4.4/4.4/4.4 & 206k/8k/2k\\
		\bottomrule
	\end{tabular}
 }
	\caption{\label{dataset_overview} Statistics of En-De and En-Ru datasets.}
\end{table}
\begin{table}[t]\centering
	\begin{threeparttable}
		\scalebox{0.8}{
			\begin{tabular}{c|cccccc}
				\toprule
				& Family  & WO &PP & GP &  GA \\
				\midrule
				English	  &IE        & SVO 	& $\None$ 		&	3SG		& SEM   \\
                Russian	  &IE        & SVO 	& $\Multiple$ 		&	3SG	& S-F  \\    
				German	  &IE        & SOV/SVO 	& $\Binary$ 		&	3SG		& S-F   \\
				Spanish	  &IE        & SVO 	& $\Binary$ 		& 1/2/3P		& SEM   \\    
				French	  &IE        & SVO 	& $\Binary$ 		&	3SG		& S-F   \\
				Japanese	  &JAP        & SOV 	& $\Avoided$ 		&	3P		& $\None$   \\
				Romanian	  &IE        & SVO 	& $\Multiple$ 		&	3SG		& S-F   \\
				Mandarin	  &ST        & SVO 	& $\Binary$ 		&	3SG		& $\None$   \\
				Vietnamese  & AA & SVO & $\Multiple$ & $\None$ & $\None$\\
    \bottomrule
			\end{tabular}
		}
	\end{threeparttable}
	\caption{\label{multi_lang_types} Properties of Languages in Our Experiments: WO (Word Order), PP (Pronouns Politeness), GP (Gendered Pronouns), and GA (Gender Assignment) denote language structural properties. IE (Indo-European), JAP (Japonic), ST (Sino-Tibetan), and AA (Austroasiatic) represent language families. Symbols $\None$, $\Binary$, $\Avoided$, and $\Multiple$ correspond to 'None', 'Binary', 'Avoided', and 'Multiple', respectively. The terms 3SG (Third Person Singular), 1/2/3P (First, Second, and Third Person), and 3P (Third Person) are used for pronoun references. SEM and S-F stand for Semantic and Semantic-Formal, respectively, in Gender Assignment.}
\end{table}

The En-Ru dataset comes from OpenSubtitle 2018 \citep{lison2018} by sampling training instances with three context sentences after tokenization and, thus, no document boundary information is preserved. In the En-De and multilingual datasets, document boundaries are provided. To maintain consistency in our translation settings during experiments, we tokenize all texts by using MeCab\footnote{\url{https://taku910.github.io/mecab/}} for Japanese, Jieba\footnote{\url{https://github.com/fxsjy/jieba}} for Chinese, VnCoreNLP \cite{vu2018} for Vietnamse and the spaCy framework\footnote{\url{https://spacy.io}} for all other languages. We also apply a sliding window with a size of $m$ sentences ($m=4$) to each document to create a similar format to that of the En-Ru dataset. 
For the first $m-1$ sentences, which do not have enough $m-1$ context sentences in the $m$-to-$m$ translation settings, we pad the beginning of these sentences with empty sentences, ensuring $m-1$ context sentences for all samples in the dataset. For preprocessing, we apply the BPE (Byte-Pair Encoding) technique from \citet{sennrich2016} with 32K merging operations to all datasets.
To identify coreference clusters for the source language, i.e., English, we leveraged the AllenNLP framework\footnote{\url{https://allenai.org}} and employed the SpanBERT large model \citep{lee2018}.
After generating sub-word units, we adjust the word-wise indices of all members in coreference clusters using the offsets for sub-word units.
\subsection{Experiment Settings}
\paragraph{Translation setting}
In our experiments,  we adopt the context-aware translation settings ($m$-to-$m$ with $m=4$) utilized in previous works \citep{zhang20}. 
For the context-agnostic setting, we translate each sentence individually.
\paragraph{Baselines systems} 
We adopt the Transformer model \citep{Vaswani17} as our two baselines: \textbf{Base Sent}, which was trained on source and target sentence pairs without context, and \textbf{Base Doc}, which was trained with contexts in the $m$-to-$m$ setting as described in \S \ref{sec_context_aware_tranformer_nmt}. 
To make a fair comparison with previous works that use similar context-aware translation settings and enhance MT system at the encoder side, we employ the \textbf{G-Transformer} \citep{bao21}, \textbf{Hybrid Context} \citep{Zheng20} and \textbf{MultiResolution} \citep{sun2022}. We also compare our approach with the \textbf{CoDoNMT} \citep{lei2022} model, which also integrates coreference resolution information to improve translation quality. Note that all aforementioned baselines utilize provided open-source code.
Additionally, we trained a simple variant of a context-aware Transformer model similar to Base Doc, but differ in that it incorporated a coreference embedding, alongside the existing positional embedding, directly in to the encoder side of the model \textbf{(Trans+C-Embedding}). This coreference embedding is derived from the original positional embedding in the encoder with the modification that all tokens within a coreference cluster share the same value as the left-most token in the same cluster.
Note that it is intended as a simple baseline for a direct model as discussed in \S \ref{sec_proposed_method}.
\begin{table*}[t]\centering
	\begin{threeparttable}
		\begin{tabular}{lccccccc}
			\toprule
			\multirow{2}[3]{*}{} & \multicolumn{3}{c}{En - Ru} & \multicolumn{3}{c}{En - De} \\
			
			\cmidrule(lr){2-4} \cmidrule(lr){5-7}
			& BL $\uparrow$ & BS $\uparrow$ &CM$\uparrow$ & BL $\uparrow$ &  BS $\uparrow$ &CM$\uparrow$  \\
			\midrule
			Base Sent 				   & 29.46 							& \textminus 9.695 		&	82.87		&  22.76  &  \textminus 6.178 & 68.06\\
			Base Doc 				   & 29.91 							& \textminus 9.551 		&	83.40		 & 21.54 & 	  \textminus 6.200 & 66.91 \\
            Hybrid Context \citep{Zheng20} 				& 29.96  						  	  &  \textminus 9.590				&	83.45     &  22.05 &\textminus 6.236   & 66.97\\
            G-Transformer \citep{bao21}				& 30.15  						  	  &  	\textminus 9.691			&	  83.13   &22.61  &  \textminus 6.090 &68.36 \\
            MultiResolution \citep{sun2022} 				&   29.85						  	  & \textminus 9.763 				&81.76	     & 22.09 &  \textminus 6.099 & 67.99\\
            DoCoNMT \citep{lei2022} 				&  29.92 						  	  &  \textminus 9.552				&	83.03     & 22.55 & \textminus 6.197  & 67.93 \\
            \midrule
			Trans+C-Embedding   &  30.13						   	   &  \textminus 9.522 			&	83.43	  & 22.54 &   \textminus 6.092 &68.80\\
            Trans+\textsc{Coref}  				  & 30.39\tnote{*}   & \textminus 9.501\tnote{$\dagger$} 	&  \textbf{83.56}\tnote{$\bullet$}  & 23.57\tnote{**}    		& \textminus 6.088\tnote{$\dagger$} & 69.17\tnote{$\diamond$}\\
			Trans+\textsc{Coref}+RR 		  & \textbf{30.43}\tnote{*}    & \textbf{\textminus 9.500}\tnote{$\dagger$} 	&  \textbf{83.56}\tnote{$\bullet$}  & \textbf{23.60}\tnote{**}  	 & \textbf{\textminus 6.086}\tnote{$\dagger$} & \textbf{69.21}\tnote{$\diamond$}\\
			\bottomrule
		\end{tabular}
            \footnotesize \tnote{(*)} and \tnote{(**)} indicate statistical significance \citep{koehn2004} at $p$ < 0.02 and $p$ < 0.01, respectively, compared to the \textit{Base Doc} system and all other baseline systems. \tnote{($\diamond$)}, \tnote{($\dagger$)}, and \tnote{($\bullet$)} signify statistical significance at $p$ < 0.05 compared to all baselines, all except Trans+C-Embedding and G-Transformer, and all except Trans+C-Embedding, Hybrid Context, and G-Transformer, respectively.
	\end{threeparttable}
	\caption{\label{overall_result} The results of all main experiments.  BL, BS and CM are abbreviations for  BLEU, BARTScore and COMET, respectively. The best performance per metric are in bold text.}
\end{table*}
\paragraph{Our systems} 
We evaluate our proposed inference methods, including the original inference method in Transformer without reranking (\textbf{Trans+\textsc{Coref}}) or with reranking with the score from our sub-model (\textbf{Trans+\textsc{Coref}+RR}) using the coreference resolution task as denoted in Equation \ref{translation_score}. 

\paragraph{Hardwares}
All models in our experiments were trained on a machine with the following specifications: an AMD EPYC 7313P CPU, 256GB RAM, a single NVIDIA RTX A6000 with 48GB VRAM, and CUDA version 11.3.  For multilingual experiments, we used a single NVIDIA RTX 3090 GPU, Intel i9-10940X, 48GB VRAM and CUDA version 12.1.
\paragraph{Hyperparameters} We use the same parameters, including the number of training epochs, learning rate, batch size, etc., for all models in our experiments. Specifically, we train all models for 40 epochs when both losses of coreference and translation in the valid set show unchanging or no improvements.  

For translation tasks, we use the Adam optimizer with $\beta_1=0.9$, $\beta_2=0.98$ and $\epsilon=1e-9$, along with an inverse square root learning rate scheduler. All \textit{dropout} values are set to 0.1, and the learning rate is set to $7e-5$. We use a batch size of 128 and 32 for experiments on the English-Russian and English-German datasets, respectively. Other parameters follow those in \citet{Vaswani17}.

For coreference tasks, we adopt parameters from \citet{kirstain21}, with some modifications to accommodate our GPU memory. We use the Adam optimizer with $\beta_1=0.9$, $\beta_2=0.98$ and $\epsilon=1e-9$, with a learning rate of $7e-5$. \textit{Dropout} value is set to 0.3,\textit{ top lambda} (the percentage of all spans to keep after filtering) is set to 0.4, \textit{hidden size} is set to 512, and the \textit{maximum span length} is set to 10. The \textit{maximum cluster value}s are set to 8 and 20 for the English-Russian and English-German datasets, respectively. To rerank the N-best translations, we use Equation \ref{translation_score} and perform a grid search on the validation set with a step size of $0.0001$ to select the optimal value for $\beta$ from $-2$ to $2$.

\subsection{Metrics}
\paragraph{BLEU}  We employ SacreBLEU \citep{post2018} as an automated evaluation metric to assess the quality of translations in our experiments.

\paragraph{BARTScore}  We follow \citet{yuan21} and use the \textit{mbart-large-50} model (\textit{mBART})\footnote{\url{https://huggingface.co/facebook/mbart-large-50}} to compute the average BARTScore of all translation to measure semantic equivalence and coherence between references and translations. In this metric, the higher value, the better semantic equivalence and coherence. 
\paragraph{COMET}  We also utilize the COMET\footnote{COMET-20 model (\textit{wmt20-COMET-da})} metric \citep{rei2020}, a neural network-based measure, since it is highly correlated to human judgement in prior work by \citet{freitag2022}.

\subsection{Results}
\begin{table*}[t]\centering
	\begin{threeparttable}
		\begin{tabular}{lccccccc}
			\toprule
			& Es $\uparrow$ & Fr $\uparrow$ &Ja$\uparrow$ & Ro $\uparrow$ &  Zh $\uparrow$ & Vi $\uparrow$ \\
			\midrule
			Base Sent 	          & 37.23 	& 37.75 		&	12.11		& 24.35  & 12.38 & 31.74 \\
			Base Doc 			  & 36.22 	& 36.89 		&	10.13		& 23.27  & 11.66 &  31.22\\
			G-Transformer 		  & 36.46  	& 37.88		& 12.27     & 24.63   & 12.07 & 32.69 \\
			Trans+\textsc{Coref}  & \textbf{38.13}\tnote{*}  & \textbf{39.01}\tnote{*}  	& \textbf{12.93}\tnote{*}   & \textbf{25.56}\tnote{*}    		& \textbf{13.18}\tnote{*} & \textbf{33.51}\tnote{*}\\
			\bottomrule
		\end{tabular}
		\begin{tablenotes}
			\footnotesize \item[*] \footnotesize With statistically significant \citep{koehn2004} at $p$ < 0.01 compared to other systems. 
		\end{tablenotes}
	\end{threeparttable}
	\caption{\label{multi_lang_result} The results of multilingual dataset in the BLEU metric. The highest results are in bold text.}
\end{table*}
The main results of our experiments are presented in Table \ref{overall_result}. Our results indicate that training the baseline Transformer model  with both context and target sentences (Base Doc) results in better performance than training with only target sentences (Base Sent) in the En-Ru dataset. This finding is consistent with those reported by \citet{voita19}, in which more contextual information is helpful to achieve better translation. However, in the En-De dataset, the Base Doc system performs worse compared to the Base Sent system. This discrepancy can be explained by the different methodologies used in constructing the En-De and En-Ru datasets. For the En-De datasets, both context-aware and context-agnostic datasets are compiled from the same pool of non-duplicate sentences. However, for the En-Ru datasets, the context-agnostic dataset is created by removing context sentences from the context-aware dataset \citep{voita19}, which results in varying numbers of non-duplicate sentences between these context-agnostic and context-aware datasets. 

When comparing our systems with the Transformer model (Base Doc), our approaches, both Trans+\textsc{Coref} and Trans+\textsc{Coref}+RR, have proven effective in enhancing translation quality by explaining the decision of translation through predicting coreference information. This is demonstrated by the superior BLEU scores ($+0.52$ in En-Ru and $+2.06$ in En-De for the Trans+\textsc{Coref}+RR),  BARTScore and COMET observed when comparing across different settings and language pairs.

Compared to the G-Transformer system described in  \citet{bao21}, our system shows an improvement in both inference approaches (Trans+\textsc{Coref} and Trans+\textsc{Coref}+RR). In the En-Ru dataset, our system achieves a higher BLEU score by $+0.24$, while in the En-De dataset, it demonstrates a larger improvement of $+1.14$ in the same metric (Trans+\textsc{Coref}). Additionally, our method outperforms the G-Transformer in terms of the BARTScore and COMET for both the En-Ru and En-De datasets. One possible explanation for these results is that the G-Transformer is specifically designed to map each sentence in the source language to only a single sentence in the target language during both training and inference steps. This design choice helps mitigating issues related to generating very long sequences. However, when the dataset size is small, as in the case of the En-De dataset, the G-Transformer encounters difficulties in selecting useful information. In contrast, our approach effectively selects useful information indirectly through the coreference explanation sub-model, especially when dealing with small-sized datasets, which allows our system to outperform under the scenarios with limited dataset size.
\begin{table*}[t!]\centering
    \scalebox{0.95}{
	    \begin{tabular}{l|p{0.75\linewidth}}
			\toprule
			Input &  \textit{ but \underline{i} 'm different . \_eos do \underline{me} just one favor . \_eos before you make any  decision ...} \textbf{\_eos meet \underline{my} team .}\\ 
			Base Doc &  \textit{{\foreignlanguage{russian}{ но \underline{я} другой . \_eos сделай \underline{мне} одолжение . \_eos прежде чем ты примешь решение ...}}} \textbf{ \foreignlanguage{russian}{\textbf{\_eos встретимся в \underline{моей} команде .}}}\\
			G-Transformer  &  \textit{\foreignlanguage{russian}{ но \underline{я} другой . \_eos сделай \underline{мне} одолжение . \_eos прежде чем ты примешь решение ...}} \textbf{ \foreignlanguage{russian}{\textbf{\_eos встреть мою команду .}}}\\
	        Trans+C-Embedding & \textit{\foreignlanguage{russian}{но \underline{я} другой . \_eos сделай \underline{мне} одолжение . \_eos прежде чем принять решение ...}} \textbf{\foreignlanguage{russian}{\_eos встретить мою команду .}}\\
			Trans+\textsc{Dec}+RR &  \textit{{\foreignlanguage{russian}{но \underline{я} другой . \_eos сделай \underline{мне} одолжение . \_eos прежде чем принять решение ...}}}  \textbf{{ \foreignlanguage{russian}{\textbf{\_eos познакомься с \underline{моей} командой .}}}}\\  
			Reference &  \textit{{\foreignlanguage{russian}{ но \underline{я} изменился . \_eos выполни только одну \underline{мою} просьбу . \_eos прежде чем ты решишь что-то ...}}} \textbf{{ \foreignlanguage{russian}{\textbf{\_eos познакомься с \underline{моей} командой .}}}}\\
			\bottomrule
		\end{tabular}
    }
\caption{\label{example_translation}Example of translations. The context and target sentences are highlighted in \textit{italics} and \textbf{bold}, respectively. Translations of the Trans+\textsc{Dec}+RR and Trans+\textsc{Dec} are identical. Underline words indicate the same mention entity.}
\end{table*}
Our method also surpasses the Transformer model with additional position embedding (Trans+C-Embedding), which relied on coreference information using a direct modeling approach. 

In the results of the multilingual TED talk dataset in Table \ref{multi_lang_result}, where we compare our proposed method to Transformer models and the best baselines in Table \ref{overall_result}, our method also surpasses other baselines within $+1.0$ to $+2.3$ BLEU scores.
These findings provide further evidence that our approach is effective in improving translation quality and can be applied to diverse language types.
\begin{table}[t]\centering
\scalebox{0.95}{
	\begin{tabular}{lcccc}
		\toprule
		& \textbf{$\alpha$} & BLEU $\uparrow$&  \\
		\midrule
		Base Sent 			& $-$	 		& 29.46    \\
		Base Doc            & $-$         & 29.91    \\ 
		\midrule
		\multirow{6}{*}{Trans+\textsc{Coref}}   & 0.8                                  & 30.36    \\
		& 1.0                                  & 30.31                               \\ 
  		& 2.0                                  & \textbf{30.39}                               \\
		& 3.0                                    &  30.27                               \\ 
		& 4.0                                    &  30.15                              \\ 
		& 10.0                                   & 30.00                               \\
		\bottomrule
	\end{tabular}
	}\caption{\label{vary_alpha_results} Ablation results on the En-Ru dataset with different weight $\alpha$. The highest result is in bold text.}
\end{table}

We provide an example of translations from our systems as well as other baseline systems in Table \ref{example_translation}. In this example, the correct translation of the phrase in the last sentence, \textit{\foreignlanguage{russian}{моей командой}} (\textit{my team}), is determined by identifying which word refers to \textit{"my"}, in this case, \textit{i} and \textit{me}. Both the G-Transformer and Trans+C-Embedding systems fail to capture these mentions and consequently produce an incorrect translation, \textit{\foreignlanguage{russian}{мою команду}}. Despite correctly translating \textit{\foreignlanguage{russian}{моей}}, the Base Doc system's phrase \textit{\foreignlanguage{russian}{встретимся в моей команде}} is grammatically incorrect and deviates from the original English \textit{"meet my team"}. 
Conversely, our systems capture this reference accurately, yielding a translation consistent with the reference.
\section{Analysis}
\label{sec_analysis}

\paragraph{Contribution of Coreference Explanation}  We conducted experiments by adjusting the value of $\alpha$ in Equation \ref{loss_fuse} during the training of the Trans+\textsc{Coref} without reranking. The experimental results in Table \ref{vary_alpha_results} indicate that for medium-sized corpora, selecting a value of $\alpha$  that is either too small or too large negatively impacts translation quality. The optimal range for $\alpha$ is $0.8 \leq \alpha \leq 2$.

\paragraph{Conditioning on Source and Target language} 
We conducted a study on coreference explanation on the En-De dataset \citep{maruf2019} with coreference cluster information as in \S \ref{section_dataset} by ablating the information from the translation so that it conditions only on the input information (Trans+\textsc{Enc}). This setting can be regarded as a conventional multi-task setting in which coreference on the input-side is predicted together with its translation. Specifically, we replace the input representation for coreference resolution sub-model from $\textsc{Dec}(\cdot)$ in Equation \ref{fuse_representation} to $\textsc{Enc}(\cdot)$  in Equation \ref{traditional_mtl} as follows
	\begin{flalign}
		\label{traditional_mtl} &\mathbf{H''}_{coref} = \Call{Enc}{\boldsymbol{H}_{enc}^{l_e}}.
	\end{flalign} 
As shown in Table \ref{transformer_with_encoder}, the conventional multi-task learning setting of Trans+\textsc{Enc} performed lower than Trans+\textsc{Coref}, which indicates the benefits of fusing information from the predicted translation.
\begin{table}[t]\centering
    \addtolength{\tabcolsep}{-0.08cm}
    \begin{threeparttable}
    \scalebox{0.95}{
	\begin{tabular}{lcccc}
		\toprule
		 &  BLEU $\uparrow$& P$\uparrow$ & R$\uparrow$ &F1$\uparrow$ \\
		\midrule
        Trans+\textsc{Enc}  & 22.98 & \textbf{85.02} & 75.69 & 80.08   \\
		\midrule
		Trans+\textsc{Coref}   & \textbf{23.57} & 82.63 & \textbf{78.31} & \textbf{80.41} \\ 
		\bottomrule
	\end{tabular}
}
\caption{\label{transformer_with_encoder} Evaluation of Trans+\textsc{Enc} and Trans+\textsc{Coref} systems using BLEU and MUC\tnote{*} metrics on the validating set of the En-De dataset. The highest results are in bold text. }
    \begin{tablenotes}
      \small
      \footnotesize \item[*] \footnotesize  The MUC metric counts the changes required to align the system's entity groupings with the gold-standard, focusing on adjustments to individual references.
    \end{tablenotes}
    \end{threeparttable}
\end{table}
\begin{figure*}[t!]
	\centering
	\includegraphics[width=11.0cm]{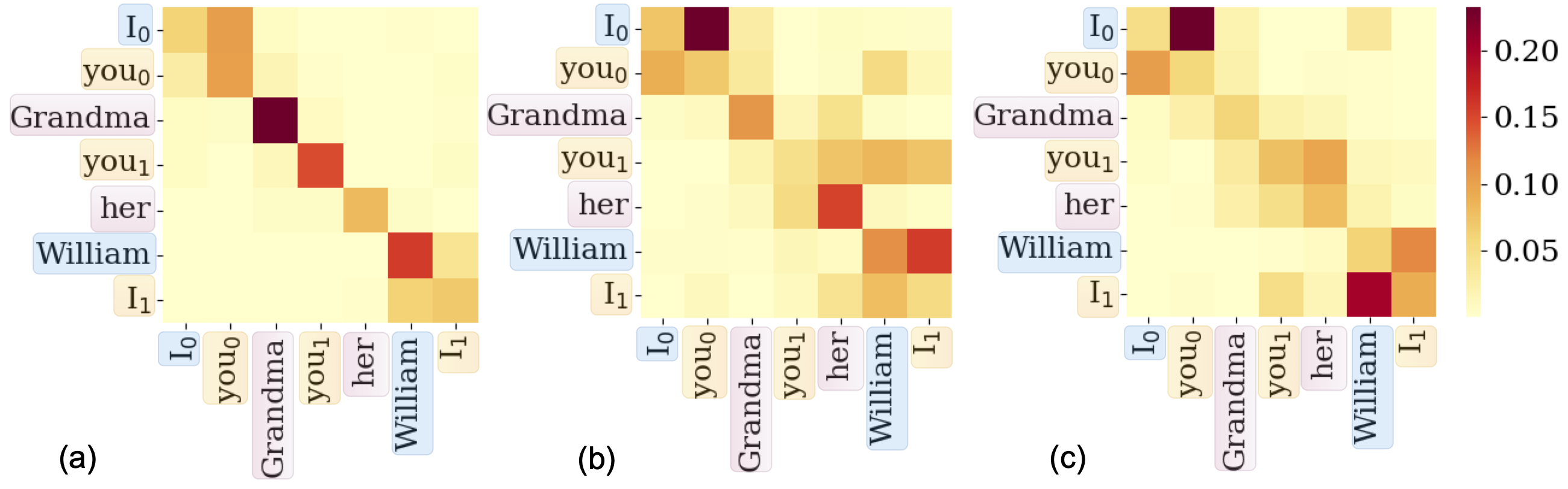}
	\caption{ Entity heat maps of self-attentions: (a) Base Doc, (b) Trans+\textsc{Enc} and (c) Trans+\textsc{Coref}.}
	\label{fig:attention_heat_map}
\end{figure*}
We further examine the entity heat maps derived from self-attention weights of both the translation and coreference sub-models in Base Doc, Trans+\textsc{Enc}, and Trans+\textsc{Coref} systems
for the input \textit{ "I\textsubscript{\_0} hate that you\textsubscript{\_0} 're leaving . Well , Grandma 's not doing well . So you\textsubscript{\_1} have to drop everything to go take care of her ?  Yes , William , I\textsubscript{\_1} do ."} from the human annotated test set from \citet{voita19}. In this particular example, the coreference clusters are defined as\textit{ [I\textsubscript{\_0}, William], [you\textsubscript{\_0}, you\textsubscript{\_1}, I\textsubscript{\_1}]}, and \textit{[Grandma, her]}. To provide visual representations, we depict the average self-attention values from the last encoder layer of these three systems. This choice is based on their tendency to focus more on semantic or abstract information \citep{clark2019}. 

Figure \ref{fig:attention_heat_map} displays the entity heat maps, which illustrate the behavior of self-attention in different systems in the translation sub-model. In the Base Doc system, self-attention primarily concentrates on local sentences while disregarding information between sentences. In contrast, the Trans+\textsc{Enc} system exhibits the ability to focus on inter-sentences. However, when it comes to tokens within coreference clusters, the focused values are incorrect for certain clusters, such as \textit{[I\textsubscript{\_0}, William]}. On the other hand, the Trans+\textsc{Coref} system not only exhibits inter-sentential focus in its self-attention heat map but also accurately depicts the focused values for tokens within coreference clusters.

\begin{figure}[t!]
	\centering
	\includegraphics[width=7.5cm]{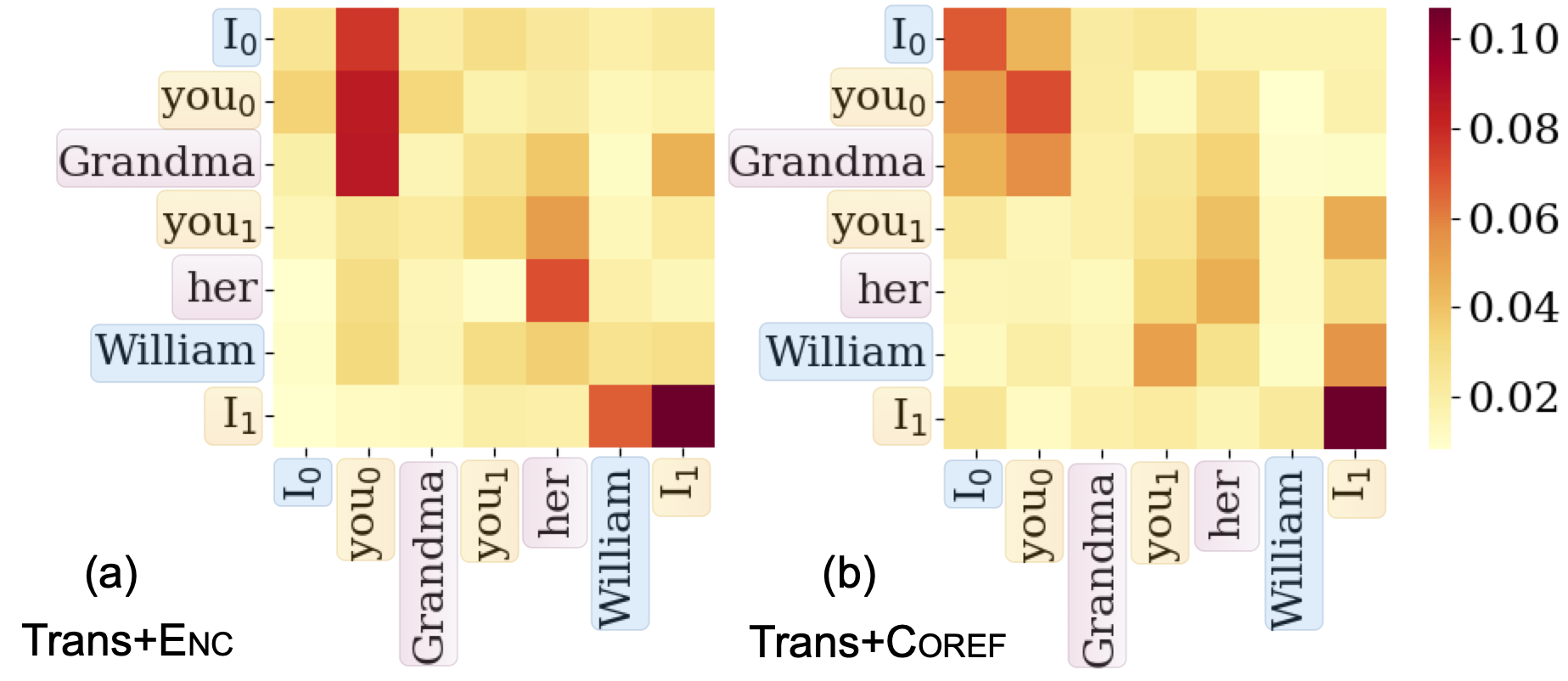}
	\caption{ Entity heat maps of self-attentions in the coreference resolution sub-model. }
	\label{fig:attention_heat_map_coref}
\end{figure} 
Figure \ref{fig:attention_heat_map_coref} demonstrates the entity heat maps in the coreference sub-model. In the Trans+\textsc{Enc} system, self-attention mainly concentrates on entities within the local scope and immediate adjacent sentences. However, when comparing these high attention values with links in the coreference clusters, a significant proportion is found to be incorrect, i.e., \textit{[Grandma; I\textsubscript{\_1}]}. On the other hand, the self-attention in Trans+\textsc{Coref} exhibits a more balanced distribution of focus across all entities within the input. This balanced distribution results in considerably fewer errors when compared to self-attention in the Trans+\textsc{Enc} system. These findings align with the MUC metric \citep{vilain95}, which is based on the minimum number of missing links in the response entities compared to the key entities, with details, particularly the F1 score, provided in Table \ref{transformer_with_encoder}. 
Note that we use reference translations to form $\boldsymbol{H}_{dec}^{l_d}$ in Equation \ref{fuse_representation} for identifying coreference clusters. Additionally, we generate gold label coreference clusters using the AllenNLP framework, as discussed in Section \ref{section_dataset}.

\paragraph{Impact of the context size}
\begin{figure}[t!]
	\centering
	\includegraphics[width=7.0cm]{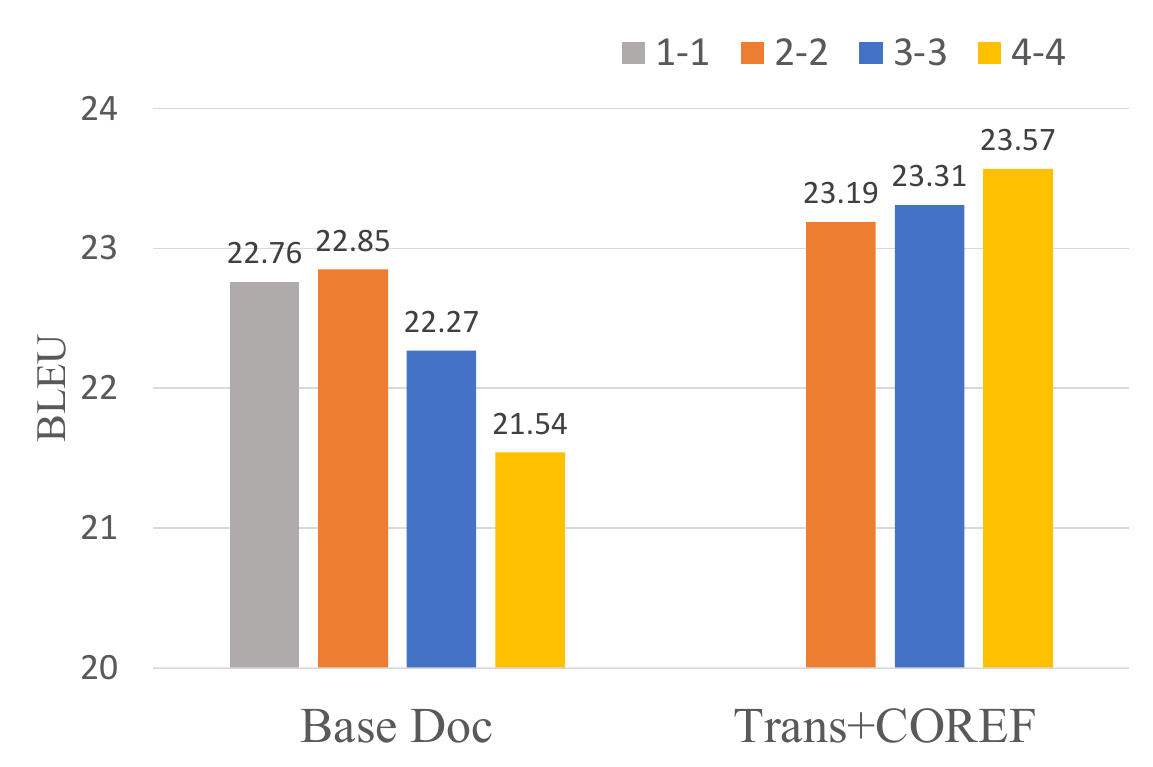}
	\caption{ Translation results on En-De datasets with different $m$-to-$m$ translation settings from $m=2$ to $m=4$. The result in the $m=1$ setting serves as the Base Sent reference. The $\alpha$ in Equation \ref{loss_fuse} is set to 4.0.  }
	\label{fig:bleu_variant_context}
\end{figure}
We conducted experiments with the \textsc{Coref} (Trans+\textsc{Coref}) and the Transformer (Base Doc) systems by exploring different context sizes in $m$-to-$m$ settings ranging from $2$ to $4$. 
The experimental results in Figure \ref{fig:bleu_variant_context} demonstrate that the Base Doc system significantly drops the translation quality when the context gets longer, while Trans+\textsc{Coref} consistently achieves gains as we incorporate more contexts.
This result also indicates the use of the coreference sub-model is able to capture contextual information better than the baseline.

\begin{table}[t]\centering
\scalebox{0.95}{
 	\begin{tabular}{lcccc}
 		\toprule
 		& \textbf{D} $\uparrow$ & \textbf{EI} $\uparrow$ & \textbf{EV} $\uparrow$  & \textbf{L} $\uparrow$ \\
 		\midrule
 		Base Doc & 83.32 & 70.20 & 62.20 & 46.0 \\
 		Trans+\textsc{Coref}  & \textbf{85.64} & \textbf{71.20} & \textbf{65.2} &  \textbf{46.4}\\
 		\bottomrule
 	\end{tabular}
 	}\caption{\label{contrastive_test_result} Experimental results on the contrastive test \citep{voita19}. D, EI, EV and L are abbreviations for Deixis, Ellipsis Infl, Ellipsis Vp and Lexical Cohesion, respectively. Note that we only utilized the text described in \S \ref{section_dataset}, while other studies may incorporate additional sentence-level bilingual and monolingual texts associated with \citet{voita19}. }
 \end{table}

\begin{figure}[t]
	\begin{tikzpicture}[scale=0.7]
		\begin{axis}[
			xtick=data,
			xticklabels from table={\RerankingResults}{firm},
			nodes near coords align={vertical},
			x=0.5cm,
			every axis plot/.append style={thick},
			ymin=30.0,
			ymax=32.0,
			ytick={30.0,30.5, 31.0},
			axis lines*=left,
			ylabel={BLEU},
			xlabel={N-best},
			legend cell align={left},
			]
			
			\addlegendentry{Trans+\textsc{Coref}+RR }
			\addplot[mark=square*,brown] table [y=value1, x expr=\coordindex,] {\RerankingResults}; %
			\addlegendentry{Trans+\textsc{Coref}}
			\addplot[mark=triangle,red] table [y=value3, x expr=\coordindex] {\RerankingResults};
			\addlegendentry{Base Doc}
			\addplot[mark=star,violet] table [y=value2, x expr=\coordindex] {\RerankingResults};
		\end{axis}
	\end{tikzpicture}
	\caption{\label{fig:rerankresult} The results with N-best variants on the En-Ru dataset \citep{voita19}.}
\end{figure}
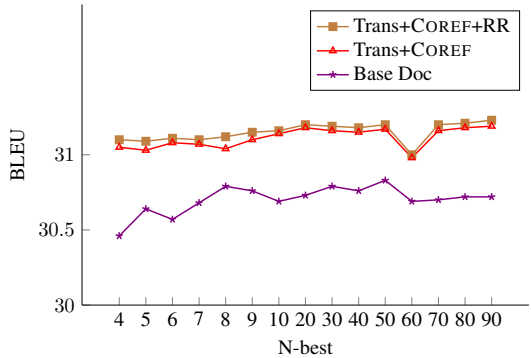
\begin{figure}[t!]
	\begin{tikzpicture}[scale=0.7]
		\begin{axis}[
			xtick=data,
			xticklabels from table={\OracleResults}{nbest},
			nodes near coords align={vertical},
			x=0.5cm,
			every axis plot/.append style={thick},
			ymin=33.0,
			ymax=48.0,
			axis lines*=left,
			ylabel={oracle BLEU},
			xlabel={N-best},
			legend cell align={left},
			]
			\addlegendentry{Trans+\textsc{Coref}}
			\addplot[mark=triangle,red] table [y=TransCoref, x expr=\coordindex] {\OracleResults};
			\addlegendentry{Base Doc}
			\addplot[mark=star,violet] table [y=BaseDoc, x expr=\coordindex] {\OracleResults};
		\end{axis}
	\end{tikzpicture}
	\caption{\label{fig:rerankresult_oracle} The results with N-best variants using the oracle BLEU metric on the En-Ru dataset \citep{voita19}.}
\end{figure}
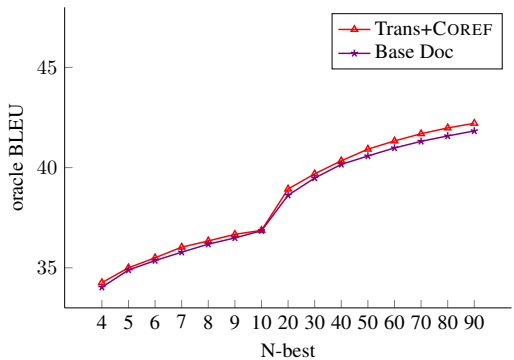
\paragraph{Impact of Coreference Explanation} We conduct experiments by reranking all translation hypotheses with varying beam sizes during inference by the Equation \ref{translation_score} to assess the impact of coreference explanation sub-model on the En-Ru dataset \citep{voita19}. Figure \ref{fig:rerankresult} illustrates the results of our experiments measured by BLEU score. 
Our findings indicate that reranking with the sub-model \textsc{Coref} yields improved results, with differences ranging from 0.02 to 0.09. 
We also report oracle BLEU score in Figure \ref{fig:rerankresult_oracle}, which is measured by selecting a hypothesis sentence that gives the maximum sentence-BLEU scores among potential hypotheses, to verify the potentially correct translations in an $N$-best list. 
The results of this experiment with differences ranging from 0.2 to 0.4 suggest that using the sub-model \textsc{Coref} has more potential to generate correct translations.
Despite the relatively minor difference in the oracle BLEU score between the Trans+\textsc{Coref} and the Base Doc systems, indicating a similarity in their candidate space, the beam search process yields better results with the Trans+\textsc{Coref} when compared with the Base Doc system. This reflects the differences in BLEU scores between Trans+\textsc{Coref} and Base Doc. The performance gap in the BLEU score between the Trans+\textsc{Coref} and Trans+\textsc{Coref}+RR could potentially be further maximized by incorporating the coreference resolution during the beam search at the expense of more computational costs. We intend to explore this possibility in our future research endeavors.

To further understand the impact of the coreference explanation sub-model on translation results, we perform an experiment on the contrastive test in \citet{voita19}, which contains human-labeled sentences to evaluate discourse phenomena and relies on the source text only, to verify whether our method can solve phenomena at the document level. 
Table \ref{contrastive_test_result}  presents the results this experiment, which indicate that our system outperforms the Base Doc system in all aspects. These results demonstrate the significant contribution of the coreference explanation sub-model to the MT system. 

\paragraph{Impact of Coreference Accuracy} We carried out experiments to assess the impact of varying accuracies within the external coreference framework, which was reported in 80.4\% of the F1 score on MUC metric for English CoNLL-2012 shared task in \citet{lee2018}, on the overall translation quality. This was achieved by randomly omitting members from coreference clusters while ensuring that each valid cluster retained a minimum of two members, i.e., removing \textit{you\textsubscript{\_1}} from the cluster \textit{ [you\textsubscript{\_0}, you\textsubscript{\_1}, I\textsubscript{\_1}]} in Figure \ref{fig:attention_heat_map}. 

Table \ref{coref_accuracy_drop} presents the outcomes of these experiments, where a slight reduction in translation quality is observed as members of coreference clusters are randomly dropped. Remarkably, even with the omission of up to half of the cluster members, the results continue to exceed the performance of the Base Doc system. This implies that our method could be robust and effective, particularly for languages with limited accuracy in coreference resolution tasks.
\begin{table}[t]\centering
    \scalebox{0.95}{
	\begin{threeparttable}
			\begin{tabular}{lccc}
				\toprule
				\multirow{2}{*}{} & \multirow{2}{*}{Pruning (\%)} & \multicolumn{2}{c}{BLEU $\uparrow$} \\ 
				&                             & \textminus RR       & $+$RR   \\ 
				\midrule
				Base Doc                                 & \textminus  & 21.54 &\textminus\\
				\midrule
				\multirow{4}{*}{Trans+\textsc{Coref}}    & 0 & 23.57 & 23.60 \\
				& 10 & 23.43 & 23.44\\
				& 20 & 23.40 & 23.41\\
				& 30 & 23.29 & 23.29\\
				& 50 & 22.86 & 22.86\\
				\bottomrule
			\end{tabular}
	\end{threeparttable}
}	\caption{\label{coref_accuracy_drop} Experimental results on dropping coreference clusters on the En-De dataset. RR means reranking with the coreference sub-model using Equation \ref{translation_score}.}
\end{table}

\begin{table}[t]\centering
\scalebox{0.95}{
	\begin{tabular}{lcc}
		\toprule
		 & Accuracy (\%) $\uparrow$&  \\
		\midrule
		Base Doc                   & 12.71   \\ 
		G-Transformer                 & 14.45    \\ 
		\midrule
		Trans+\textsc{Coref} & 18.50\\
		\bottomrule
	\end{tabular}
 }
	\caption{\label{deixis_we} Accuracy of translating the word \textit{we} into Vietnamese (173 samples).
	}
\end{table}

\begin{figure}[t]
	\begin{tikzpicture}[scale=0.7]
		\begin{axis}[
			xtick=data,
			nodes near coords align={vertical},
			every axis plot/.append style={thick},
			ytick={25, 26, 27, 28, 29, 30,  31},
			legend cell align=left,
			axis lines*=left,
			ymin=25.0,
			ymax=32.5,
			ylabel={BLEU},
			xlabel={Corpus size (in millions)},
		]
			
			\addlegendentry{Base Doc}
			\addplot[mark=star,violet] coordinates {
				(0.2,25.59)
				(0.5,28.47)
				(1,29.32)
				(1.5,29.91)
				
			};
			\addlegendentry{Trans+\textsc{Coref}}
			\addplot[mark=triangle,red] coordinates {
				(0.2,26.15)
				(0.5,28.78)
				(1,29.74)
				(1.5,30.39)
			};
			
		\end{axis}
	\end{tikzpicture}
\caption{ Translation results on the En-Ru dataset \citep{voita19} with different sample sizes.}
\label{fig:bleu_size}
\end{figure}
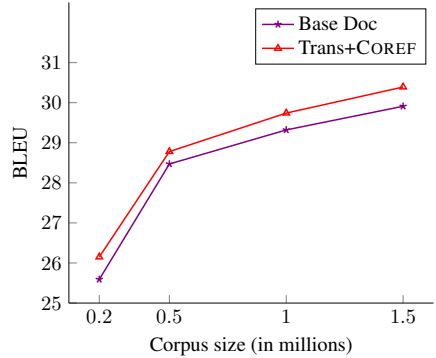

\paragraph{Impact of the corpus size}

We randomly sampled training instances from the En-Ru dataset and varied the sample sizes to 200,000 (comparable size to the En-De dataset), 500,000, and 1,000,000. Subsequently, we evaluated the contribution of the \textsc{Coref} sub-model (Trans+\textsc{Coref}) and the Transformer (Base Doc) on these datasets of different sample sizes. Figure \ref{fig:bleu_size} illustrates the results of these experiments. 
Our proposed system outperforms the Transformer model (Base Doc) across all sample sizes in the test set. Notably, this improvement is not limited to the small dataset size setting but similar trends are observed for medium-sized datasets. These results indicate that our system consistently outperforms the transformer model and achieves improved translation qualities regardless of the dataset size.

\paragraph{Remaining Challenges and Unresolved Questions} 
While our proposed method and existing works enhance translation accuracy for certain linguistic phenomena, challenges persist, particularly in handling deixis. Unlike straightforward scenarios where additional context aids in accurately translating deictic terms (e.g., determining the speakers in a conversation to correctly translate the words \textit{I} and \textit{You}), some instances require a comprehensive understanding of the provided text's content to achieve correct pronoun translation.    
Consider the following example from the test data of the English-Vietnamese dataset \citep{qi2018}: \textit{\foreignlanguage{vietnam}{"Oh my god! you're right! who can \(we_{\text{[chúng ta]}}\) sue? Now Chris is a really brilliant lawyer, but he knew almost nothing about patent law and certainly nothing about genetics. I knew something about genetics, but I wasn't even a lawyer, let alone a patent lawyer. So clearly \(we_{\text{[chúng tôi]}}\) had a lot to learn before \(we_{\text{[chúng ta]}}\) could file a lawsuit."}} In this context, the English word \textit{we} is translated as either \textit{\foreignlanguage{vietnam}{chúng tôi} (\(we_{\text{[chúng tôi]}}\))} or \textit{\foreignlanguage{vietnam}{chúng ta} (\(we_{\text{[chúng ta]}}\))}, reflecting the exclusion or inclusion of the listener. This example underscores the importance of contextual nuances in translating pronouns like \textit{we} or \textit{us} from English to Vietnamese, where the choice between \textit{chúng tôi} and \textit{chúng ta} is critical. 

Building on the insights from the described example, we extracted all samples that presented similar linguistic challenges, in which a correctly translated sample must ensure that every instance of the word \textit{we} is accurately translated. Table \ref{deixis_we} presents the accuracy of translating the word \textit{we} into the correct Vietnamese. While our method surpasses other baseline models in performance, it still exhibits lower accuracy in comparison to the deixis-related outcomes of the contrastive test for Russian  \citep{voita19}. This discrepancy highlights the phenomenon as a significant challenge that warrants further investigation.

\paragraph{Computational Cost} We present a detailed comparison of the parameter count and training time per epoch for our proposed method alongside other baselines in Table \ref{parameters_of_models}.  When compared to the G-Transformer, our method uses fewer parameters, takes less time to train, and yet achieves better performance. On the other hand, the Base Doc system uses the fewest parameters and trains the quickest, but its results are notably underperforming.

\begin{table}[t]\centering
\scalebox{0.95}{
 	\begin{tabular}{lccc}
 		\toprule
 		& \parbox[c]{1.1cm}{\centering No. of \\ Params} & \parbox[c]{1.0cm}{\centering Training \\ Time} & \parbox[c]{1.3cm}{\centering En-De $\uparrow$}\\
 		\midrule
 		Base Doc & 92.03 & 407 & 21.54  \\
   MultiResolution & 92.03 & 610 & 22.09  \\
        G-Transformer & 101.48 & 566 & 22.61 \\
        Hybrid Context & 65.78 & 1,776  & 22.05 \\
        CoDoNMT & 92.03 & 638  & 22.55  \\
        Trans+\textsc{Coref} & 98.59 & 503 & 23.57 \\
 		\bottomrule
 	\end{tabular}
 	}\caption{\label{parameters_of_models} Number of parameters (in million), training time for one epoch (in seconds) and results of systems (in the BLEU metric) on the En-De dataset.}
 \end{table}
\section{Related Works}
\label{sec_related_works}
Multi-task learning has primarily been utilized in MT tasks to integrate external knowledge into MT systems. \citet{luong16, niehues2017, eriguchi2017} have employed multi-task learning with different variations of shared weights of encoders, decoders, or attentions between tasks to effectively incorporate parsing knowledge into sequence-to-sequence MT systems.

For incorporating coreference cluster information, \citet{ohtani2019}, \citet{xu2021} and \citet{lei2022} incorporate coreference cluster information to improve their NMT models. \citet{ohtani2019} integrates coreference cluster information  into a graph-based NMT approach to enhance the information. Similarly, \citet{xu2021}  uses the information to connect words across different sentences and incorporates other parsing information to construct a graph at the document-level, resulting in an improvement in translation quality. \citet{lei2022} employs coreference information to construct cohesion maskings and fune-tunes sentence MT systems to produce more cohesive outputs.
On the other hand, \citet{ stojanovski2018, Hwang21}  leverage coreference cluster information through augmented steps. They either add noise to construct a coreference-augmented dataset or use coreference information to create a contrastive dataset and train their MT systems on these enhanced datasets to achieve better translation performance. 
For context-aware MT, \citet{kuang17} and \citet{tu2018} focus on utilizing memory-augmented neural networks, which store and retrieve previously translated parts in NMT systems. These approaches help unify the translation of objects, names, and other elements across different sentences in a paragraph. In contrast, \citet{xiong19, voita19} develop a multiple-pass decoding method inspired by the Deliberation Network \cite{Xia17} to address coherence issues, i.e., deixis and ellipsis in paragraphs. They first translate the source sentences in the first pass and then correct the translations to improve coherence in the second pass. \citet{Mansimov20} introduce a self-training technique, similar to domain self-adaptation, to develop a document-level NMT system. Meanwhile, various methods aim to encapsulate contextual information, i.e., hierachical attention \citep{maruf2019}, multiple-attention mechanism \citep{zhang20, bao21}, recurrent memory unit \citep{feng2022} \footnote{In \citet{feng2022}, they provided source code without instructions. We tried to reuse and reimplement their method; however, we can not reproduce their results in any efforts. They did not reply our emails for asking training details. We therefore decide not to include their results in Table \ref{overall_result}.}. In a data augmentation approach, \citet{bao2023} diversify training data for the target side language, rather than only using a single human translation for each source document.

Recently, \citet{wang2023} has shown that state-of-the-art Large Language Models (LLMs), i.e. GPT-4 \citep{openai2024gpt4}, outperform traditional translation models in context-aware MT. In other approaches, \citet{wu2024} and \citet{li2024} have developed effective fine-tuning and translation methods for lightweight LLMs; however, the efficacy of NMT models can exceed that of lightweight LLMs, varying by language pair.

\section{Conclusion}
\label{sec_conclusion}
This study presents a context-aware MT model that explains the translation output by predicting coreference clusters in the source side. The model comprises two sub-models, a translation sub-model and a coreference resolution sub-model, with no modifications to the translation model. The coreference resolution sub-model predicts coreference clusters by fusing the representation from both the encoder and decoder to capture relations in the two languages explicitly. Under the same settings of the En-Ru, En-De and the multilingual datasets, and following analyses on the coreference sub-model’s contributions, the impacts of context and corpus size, as well as the type of information utilized in the sub-model, our proposed method has proven effective in enhancing translation quality.
\section*{Limitations}
In this study, the hidden dimension size in the coreference resolution sub-model is smaller than typical state-of-the-art systems, i.e., 512 vs. 2048, potentially limiting its accuracy and negatively impacting the quality of translation. Additionally, this study requires fine-tuning for a certain hyperparameter that combines the coreference resolution sub-model and the translation model to achieve satisfactory results.
\section*{Acknowledgements}
The authors are grateful to the anonymous reviewers and the Action Editor who provided many insightful comments that improve the paper.
This work was supported by JSPS KAKENHI Grant Number JP21H05054.


\bibliographystyle{acl_natbib}
\bibliography{tacl2021.bib}
\iftaclpubformat







  
\fi

\end{document}